\documentclass[10pt,twocolumn,letterpaper]{article}
\pdfoutput=1

\usepackage{cvpr}
\usepackage{times}
\usepackage{epsfig}
\usepackage{graphicx}
\usepackage{amsmath}
\usepackage{amssymb}
\usepackage{bm}
\usepackage{booktabs}
\usepackage{color}
\usepackage{multirow}
\usepackage{subfigure}
\usepackage{tabu}

\usepackage[pagebackref=true,breaklinks=true,letterpaper=true,colorlinks,bookmarks=false]{hyperref}

\cvprfinalcopy

\begin{document}

\title{Contrastive Learning for Compact Single Image Dehazing}

\renewcommand{\thefootnote}{\fnsymbol{footnote}}
\author {
	Haiyan Wu \textsuperscript{\rm 1}\footnotemark[1],
	\quad Yanyun Qu \textsuperscript{\rm 2} \footnotemark[1],
	\quad Shaohui Lin \textsuperscript{\rm 1}\footnotemark[2]
	\quad Jian Zhou \textsuperscript{\rm 3}, \\
	\quad Ruizhi Qiao \textsuperscript{\rm 3}, 
	\quad Zhizhong Zhang \textsuperscript{\rm 1},
	\quad Yuan Xie \textsuperscript{\rm 1}\footnotemark[2],
	\quad Lizhuang Ma \textsuperscript{\rm 1}, \\
	\textsuperscript{\rm 1}School of Computer Science and Technology, East China Normal University, Shanghai, China\\
	\textsuperscript{\rm 2}School of Information Science and Engineering, Xiamen University, Fujian, China \\
	\textsuperscript{\rm 3}Tencent Youtu Lab, Shanghai, China\\
	{\tt\small 51194501183@stu.ecnu.edu.cn, yyqu@xmu.edu.cn,
	}\\
	{\tt\small \{shlin,yxie,zzzhang,lzma\}@cs.ecnu.edu.cn, \{darnellzhou,ruizhiqiao\}@tencent.com}
}

\maketitle

\footnotetext[1]{Equal contribution.}
\footnotetext[2]{Corresponding author.}
\renewcommand{\thefootnote}{\arabic{footnote}}

\thispagestyle{empty}

\begin{abstract}
Single image dehazing is a challenging ill-posed problem due to the severe information degeneration. However, existing deep learning based dehazing methods only adopt clear images as positive samples to guide the training of dehazing network while negative information is unexploited. Moreover, most of them focus on strengthening the dehazing network with an increase of depth and width, leading to a significant requirement of computation and memory. 
In this paper, we propose a novel contrastive regularization (CR) built upon contrastive learning to  exploit both the information of hazy images and clear images as negative and positive samples, respectively. CR ensures that the restored image is pulled to closer to the clear image and pushed to far away from the hazy image in the representation space. 

Furthermore, considering trade-off between performance and memory storage, we develop a compact dehazing network based on autoencoder-like (AE) framework. It involves an adaptive mixup operation and a dynamic feature enhancement module, which can benefit from preserving information flow adaptively and expanding the receptive field to improve the network's transformation capability, respectively. We term our dehazing network with autoencoder and contrastive regularization as AECR-Net. 
The extensive experiments on synthetic and real-world datasets demonstrate that our AECR-Net surpass the state-of-the-art approaches. The code is released in \url{https://github.com/GlassyWu/AECR-Net}.
\end{abstract}

\begin{figure}[t]
\centering
\vspace{-1em}
  \subfigure[Hazy input]{
    \includegraphics[scale = 0.24]{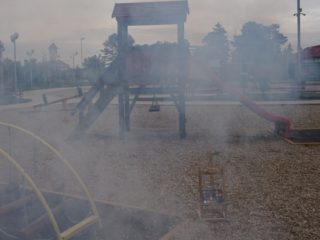}
    \label{Fig:intro_a}
    }
    \hspace{-1em}
  \subfigure[Only L1 loss \cite{qin2020ffa}]{
    \includegraphics[scale = 0.24]{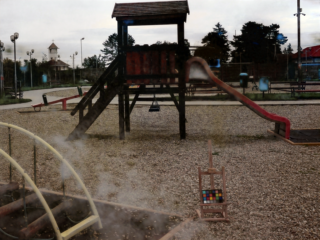}
    \label{Fig:intro_b}
  }
  \hspace{-1em}
  \subfigure[Prior \cite{Shao_2020_CVPR}]{
    \includegraphics[scale = 0.24]{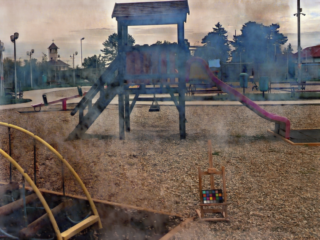}
    \label{Fig:intro_c}
  }
  \subfigure[KDDN \cite{hong2020distilling}]{
    \includegraphics[scale = 0.24]{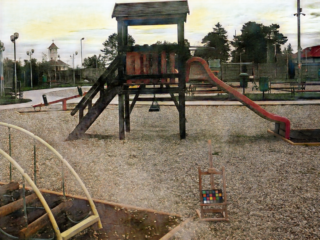}
    \label{Fig:intro_d}
  }
  \hspace{-1em}
  \subfigure[Our CR]{
    \includegraphics[scale = 0.24]{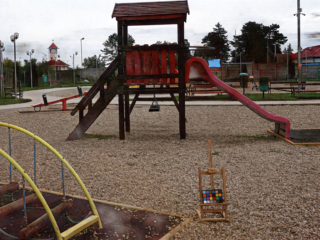}
    \label{Fig:intro_e}
  }
  \hspace{-1em}
  \subfigure[Ground-truth]{
    \includegraphics[scale = 0.24]{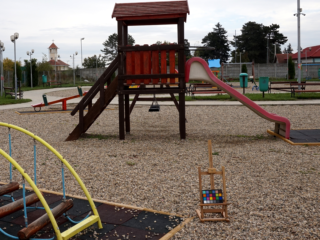}
    \label{Fig:intro_f}
  }
 \caption{Comparison with only positive-orient supervision.} 
 \label{Fig:intro}
\end{figure}

\section{Introduction}
Haze is an important factor to cause noticeable visual quality degradation in object appearance and contrast. Input images captured under hazy scenes significantly affect the performance of high-level computer vision tasks, such as object detection \cite{li2017end,Chen_2018_CVPR} and scene understanding \cite{Sakaridis_2018_ECCV, sakaridis2018semantic}. Therefore, image dehazing has received a great deal of research focus on image restoration for helping to develop effective computer vision systems. 

\begin{figure}[t]
	\centering
	\includegraphics[width=8cm]{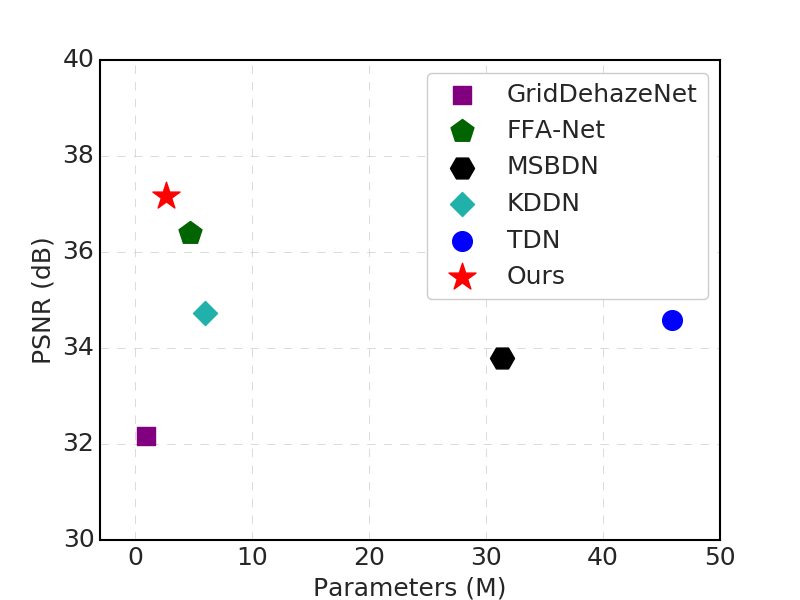}
	\caption{The best PSNR-parameter trade-off of our method.}
	\label{Fig:trade-off}
	\vspace{-0.2cm}
\end{figure}

Recently, various end-to-end CNN-based methods \cite{Qu_2019_CVPR, liuICCV2019GridDehazeNet, qin2020ffa, hong2020distilling, Dong_2020_CVPR, Shao_2020_CVPR} have been proposed to simplify the dehazing problem by directly learning hazy-to-clear image translation via a dehazing network. 
However, there exists several issues:
(1) \emph{Less effectiveness of only positive-orient dehazing objective function}. 
Most existing methods \cite{cai2016dehazenet, li2017aod,qin2020ffa,Dong_2020_CVPR} typically adopt clear images (\emph{a.k.a. ground-truth}) as positive samples\footnote{In this paper, positive samples, clear images and ground-truth are the same concept in the image dehazing task.} to guide the training of dehazing network via L1/L2 based image reconstruction loss without any regularization. However, only image reconstruction loss is unable to effectively deal with the details of images, which may lead to color distortion in the restored images (see Fig. \ref{Fig:intro_b}). Recently, additional knowledge from positive samples based regularization \cite{hong2020distilling, Shao_2020_CVPR,zhang2018densely,liuICCV2019GridDehazeNet} has been proposed to make the dehazing model generate more natural restored images. For example, Hong \etal \cite{hong2020distilling} introduced an additional teacher network to transfer knowledge from the intermediate representation of the positive image extracted by the teacher to the student/dehazing network as positive samples based regularization. 
Although they utilize the information of positive images as an upper bound, the artifacts or unsatisfied results still happen due to the unexploited information of negative images as an lower bound (see Fig. \ref{Fig:intro_d}).
(2) \emph{Parameter-heavy dehazing networks}. 
Previous works \cite{liuICCV2019GridDehazeNet,guo2019dense,qin2020ffa,Dong_2020_CVPR,Liu_2020_CVPR_Workshops} focus on improving the dehazing performance by significantly increasing the depth or width of the dehazing models without considering memory or computation overhead, which prohibits their usage on resource-limited environments, such as mobile or embedded devices. For example, TDN \cite{Liu_2020_CVPR_Workshops}, the champion model on NTIRE 2020 Challenge \cite{Ancuti_2020_CVPR_Workshops} in the dehazing task has 46.18 million parameters. More state-of-the-art (SOTA) models about their performance and parameters are presented in Fig. \ref{Fig:trade-off}.

To address these issues, we propose a novel contrastive regularization (CR), which is inspired by contrastive learning \cite{hadsell2006dimensionality,oord2018representation,henaff2020data,he2020momentum,chen2020simple}. 

\begin{figure*}[t]
	\centering
	\includegraphics[width=17cm]{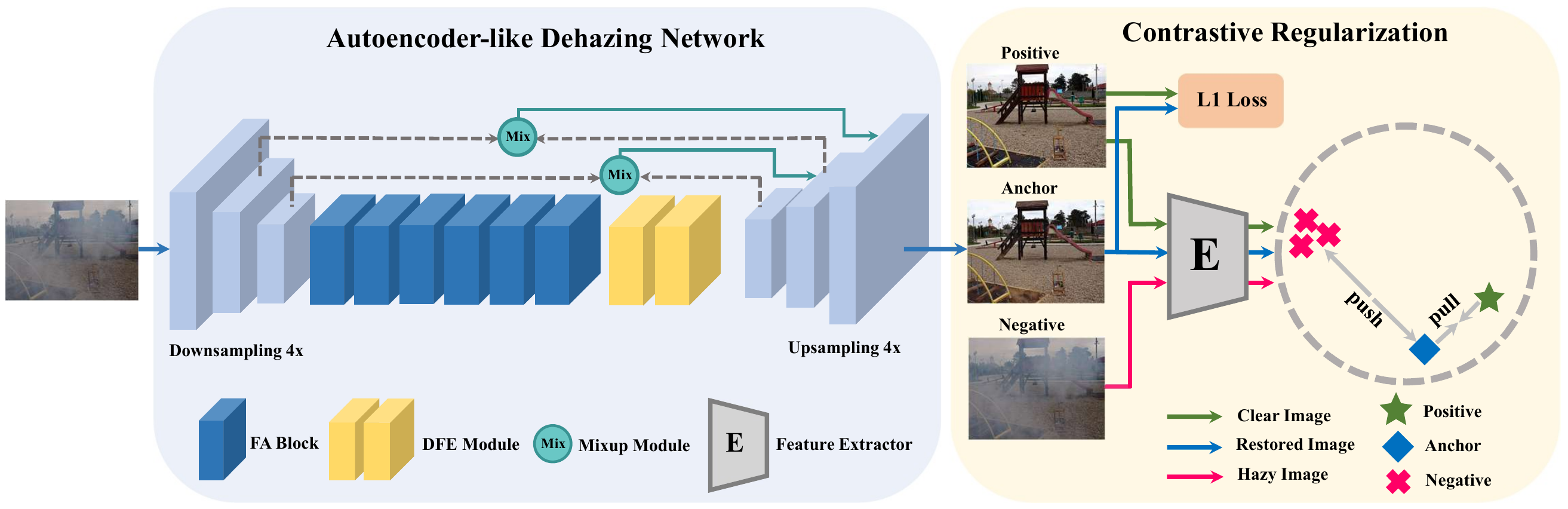}
	\caption{The architecture of the proposed AECR-Net. It consists of autoencoder-like (AE) dehazing network and constrative regularization (CR). AE has light parameters with one 4$\times$ downsampling module, six FA blocks, one DFE module, one $4\times$ upsampling module and two adaptive mixup operations. We jointly minimize the L1 based reconstruction loss and constrative regularization to better pull the restored image (\ie anchor) to the clear (\ie positive) image and push the restored image to the hazy (\ie negative) image.}
	\label{Fig:CDNet}
	\vspace{-0.2cm}
\end{figure*}

As shown in the right panel of Fig. \ref{Fig:CDNet}, we denote a hazy image, its corresponding restored image generated by a dehazing network and its clear image (\ie ground-truth) as negative, anchor and positive respectively. There are two “opposing forces”; One pulls the prediction closer to the clear image, the other one pushes the prediction farther away from the hazy image in the representation space. 
Therefore, CR constrains the anchor images into the closed upper and lower bounds via contrastive learning, which better help the dehazing network approximate the positive images and move away from the negative images.
Furthermore, CR improves the performance for image dehazing without introducing additional computation/parameters during testing phase, since it can be directly removed for inference. 

To achieve the best trade-off between performance and parameters, we also develop a compact dehazing network by adopting autoencoder-like (AE) framework to make dense convolution computation in the low-resolution space and also reduce the number of layers, which is presented in Fig. \ref{Fig:CDNet}. The information loss from the reduction of parameters can be made up by \emph{adaptive mixup} and \emph{dynamic feature enhancement} (DFE). Adaptive mixup enables the information of shallow features from the downsampling part adaptively flow to high-level features from the upsampling one, which is effective for feature preserving. Inspired by deformable convolution \cite{zhu2018deformable} with strong transformation modeling capability, DFE module dynamically expands the receptive field for fusing more spatially structured information, which significantly improves the performance of our dehazing network.
We term the proposed image dehazing framework  as AECR-Net by leveraging contrastive regularization into the proposed AE-like dehazing network.
 
Our main contributions are summarized as follows:
\begin{itemize}
	\item We propose a novel ACER-Net to effectively generate high quality haze-free images by contrastive regularization and highly compact autoencoder-like based dehazing network. AECR-Net achieves the best parameter-performance trade-off, compared to the state-of-the-art approaches.
	\item The proposed contrastive regularization as a universal regularization can further improve the performance of various state-of-the-art dehazing networks.
	\item Adaptive mixup and dynamic feature enhancement module in the proposed autoencoder-like (AE) dehazing network can help the dehazing model preserve information flow adaptively and enhance the network's transformation capability, respectively. 
\end{itemize} 

\section{Related Work}

\subsection{Single Image Haze Removal}
Single image dehazing aims to generate the haze-free images from the hazy observation images, which can be categorized into prior-based methods \cite{tan2008visibility,he2010single,zhu2014single,berman2016non} and learning-based methods \cite{cai2016dehazenet, li2017aod, zhang2018densely,Qu_2019_CVPR,hong2020distilling}. 

\textbf{Prior-based Image Dehazing Methods.} 
These methods depend on the physical scattering model \cite{mccartney1976optics} and usually remove the haze using handcraft priors from empirical observation, such as contrast maximization \cite{tan2008visibility}, dark channel prior (DCP) \cite{he2010single}, color attenuation prior \cite{zhu2014single} and non-local prior \cite{berman2016non}.
Although these prior-based methods achieve promising results, the priors depend on the relative assumption and specific target scene, which leads to less robustness in the complex practical scene. For instance, DCP \cite{he2010single} cannot well dehaze the sky regions, since it does not satisfy with the prior assumption.

\textbf{Learning-based Image Dehazing Methods.} 
Different from prior-based methods, learning-based methods are data-driven, which often use deep neural networks to estimate the transmission map and atmospheric light in the physical scattering model \cite{cai2016dehazenet,ren2016single,li2017aod,zhang2018densely} or directly learn hazy-to-clear image translation \cite{ren2018gated,liuICCV2019GridDehazeNet,Qu_2019_CVPR,qin2020ffa,Dong_2020_CVPR}. 

Early works \cite{cai2016dehazenet,ren2016single,li2017aod,zhang2018densely} focus on directly estimating the transmission map and atmospheric light. However, these methods may cause a cumulative error to generate the artifacts, since the inaccurate estimation or some estimation bias on the transmission map and the global atmospheric light results in large reconstruction error between the restored images and the clear ones. Besides, it is difficult or expensive to collect the ground-truth about transmission map and global atmospheric light in the real world.

Recently, various end-to-end methods \cite{ren2018gated,chen2019gated,Qu_2019_CVPR,liuICCV2019GridDehazeNet,hong2020distilling,qin2020ffa,Dong_2020_CVPR} have been proposed to directly learn hazy-to-clear image translation without using atmospheric scattering model.
Most of them \cite{ren2018gated,liuICCV2019GridDehazeNet,qin2020ffa,Dong_2020_CVPR} focus on strengthening the dehazing network and adopt clear images as positive samples to guide the dehazing network via image reconstruction loss without any regularization on images or features. For instance, Qin \etal \cite{qin2020ffa} proposed a feature fusion attention mechanism network to enhance flexibility by dealing with different types of information, which only uses L1 based reconstruction loss between the restored image and ground-truth. 
Dong \etal \cite{Dong_2020_CVPR} proposed a boosted decoder to progressively restore the haze-free image by only considering the reconstruction error using ground-truth as supervision. To better use the knowledge from positive samples, Hong \etal \cite{hong2020distilling} introduced an additional teacher network to transfer knowledge from the intermediate representation of the positive image extracted by the teacher to the student/dehazing network. 
Although these methods utilize the information of positive images as an upper bound, the artifacts or unsatisfied results still happen due to the unexploited information of negative images as an lower bound. Moreover, these methods are also performance-oriented to significantly increase the depth of the dehazing network, which leads to heavy computation and parameter costs.

Different from these methods, we propose a novel contrastive regularization to exploit both the information of negative images and positive images via contrastive learning. Furthermore, our dehazing network is compact by reducing the number of layers and spatial size based on autoencoder-like framework.

\subsection{Contrastive Learning}
Contrastive learning are widely used in self-supervised representation learning \cite{henaff2019data,tian2019contrastive,sermanet2018time,he2020momentum,chen2020simple}, where the contrastive losses are inspired by noise contrastive estimation \cite{gutmann2010noise}, triplet loss \cite{hermans2017defense} or N-pair loss \cite{sohn2016improved}. 
For a given anchor point, contrastive learning aims to pull the anchor close to positive points and push the anchor far away from negative points in the representation space. 
Previous works \cite{chen2020simple, he2020momentum, henaff2020data,grill2020bootstrap} often apply contrastive learning into high-level vision tasks, since these tasks inherently suit for modeling the contrast between positive and negative samples/features. 
Recently, the work in \cite{park2020contrastive} has demonstrated that contrastive learning can improve unpaired image-to-image translation quality. 
However, there are still few works to apply constrative learning into image dehazing, as the speciality of this task on constructing contrastive samples and contrastive loss. Moreover, different from \cite{park2020contrastive}, we proposed a new sampling method and a novel pixel-wise contrastive loss (\emph{a.k.a.} contrastive regularization).

\section{Our Method}
In this section, we first describe the notations.  Then, we present the proposed autoencoder-like (AE) dehazing network using adaptive mixup for better feature preserving and a dynamic feature enhancement module for fusing more spatially structured information. Finally, we employ contrastive regularization as a universal regularization applied into our AE-like dehazing network.

\subsection{Notations}
End-to-end single image dehazing methods \cite{Qu_2019_CVPR, liuICCV2019GridDehazeNet, hong2020distilling, Shao_2020_CVPR} remove haze images by using two losses, image reconstruction loss and regularization term on the restored image, which can be formulated as:
\begin{equation}
\arg\min_{w}\|J-\phi(I,w)\|+\beta\rho(\phi(I,w)),
\label{loss_IR}
\end{equation}
where $I$ is a hazy image, $J$ is the corresponding clear image, and $\phi(\cdot,\theta)$ is the dehazing network with parameter $w$. $\|J-\phi(I,w)\|$ is the data fidelity term, which often uses L1/L2 norm based loss. $\rho(\cdot)$ is the regularization term to generate a nature and smooth dehazing image, where TV-norm \cite{Shao_2020_CVPR, 8902220}, DCP prior \cite{Shao_2020_CVPR, 8902220} are widely used in the regularization term. $\beta$ is a penalty parameter for balancing the data fidelity term and regularization term. Different from the previous regularization, we employ a contrastive regularization to improve the quality of the restored images.

\subsection{Autoencoder-like Dehazing Network.}
Inspired by FFA-Net \cite{qin2020ffa} with high effective FA blocks, we use the FA block as our basic block in the proposed autoencoder-like (AE) network. 
Different from FFA-Net, we significantly reduce the memory storage to generate a compact dehazing model. 
As presented in Fig. \ref{Fig:CDNet}, the AE-like network first adopts 4$\times$ downsampling operation (\eg one regular convolution with stride 1 and two convolution layers all with stride 2) to make dense FA blocks learn the feature representation in the low-resolution space, and then employ the corresponding 4$\times$ upsampling and one regular convolution to generate the restored image. Note that we significantly reduce the number of FA blocks by only using 6 FA blocks (\emph{vs.} 57 FA blocks in FFA-Net). 
To improve the information
flow between layers and fuse more spatially structured information, we propose two different connectivity
patterns: (1) Adaptive mixup dynamically fuses the features between the downsampling layers and the upsampling layers for feature preserving. (2) Dynamic feature enhancement (DFE) module enhances the transformation capability by fusing more spatially structured information.  

\begin{figure}[t]
	\centering
	\includegraphics[width=8cm]{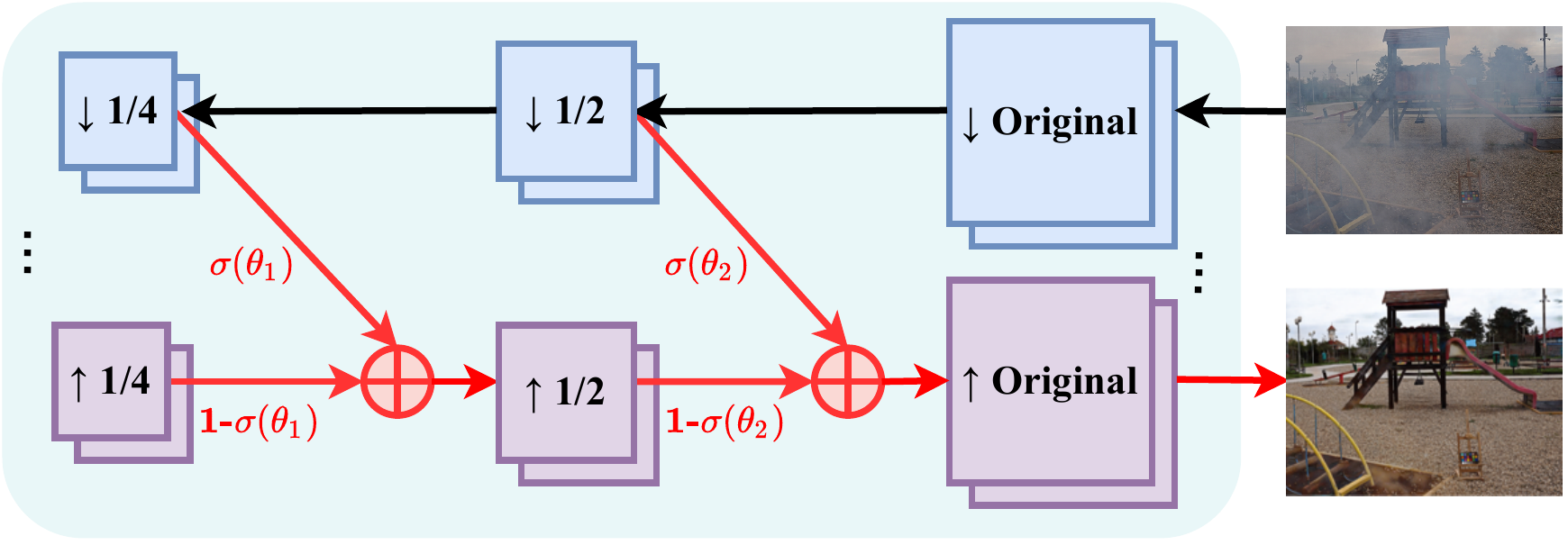}
	\caption{Adaptive mixup. The first and second rows are downsampling and upsampling operations, respectively.}
	\label{Fig:Mix}
	\vspace{-0.2cm}
\end{figure}

\subsubsection{Adaptive Mixup for Feature Preserving}
\label{Mix}
Low-level features ($\eg$ edges and contours) can be captured in the shallow layers of CNNs \cite{zeiler2014visualizing}.
However, with an increase of the network's depth, the shallow features degrades gradually \cite{he2016deep}. 
To deal with this issue, several previous works \cite{ronneberger2015u, he2016deep} integrate the shallow and deep features to generate new features via the skip connections with an addition or concatenation operation. 
Actually, FA block \cite{qin2020ffa} also use addition based skip connections to fuse the internal input and output features. 
However, there are missing connection between the features from the downsampling layers and upsampling layers in our image dehazing network, which causes shallow features (\eg edge and corner) lost. Thus, we apply the adaptive mixup operation \cite{zhang2017mixup} to fuse the information from these two layers for feature preserving (see Fig. \ref{Fig:Mix}).
In our case, we consider two downsampling layers and two upsampling layers, such that the final output of the mixup operations can be formulated as: 
\begin{equation}
\small
\label{skip_connection}
\begin{split}
\bm{f}_{\uparrow2}&=\text{Mix}(\bm{f}_{\downarrow1}, \bm{f}_{\uparrow1}) = \sigma (\theta_1)*\bm{f}_{\downarrow1} + (1-\sigma (\theta_1))*\bm{f}_{\uparrow1},\\
\bm{f}_{\uparrow}&=\text{Mix}(\bm{f}_{\downarrow2}, \bm{f}_{\uparrow2}) = \sigma (\theta_2)*\bm{f}_{\downarrow2} + (1-\sigma (\theta_2))*\bm{f}_{\uparrow2},\\
\end{split}
\end{equation}

where $\bm{f}_{\downarrow i}$ and $\bm{f}_{\uparrow i}$ are feature maps from the $i$-th downsampling and upsampling layer, respectively. $\bm{f}_{\uparrow}$ is the final output. 
$\sigma(\theta_i), i=1,2$ is the $i$-th learnable factor to fuse the inputs from the $i$-th downsampling layer and the $i$-th upsampling one, whose value is determined by the sigmoid operator $\sigma$ on parameter $\theta_i$. During training, we can effectively learn these two learnable factors, which achieves better performance than the constant factors (see Section \ref{ablation_study}).

\begin{figure}[t]
	\centering
	\includegraphics[width=8.5cm]{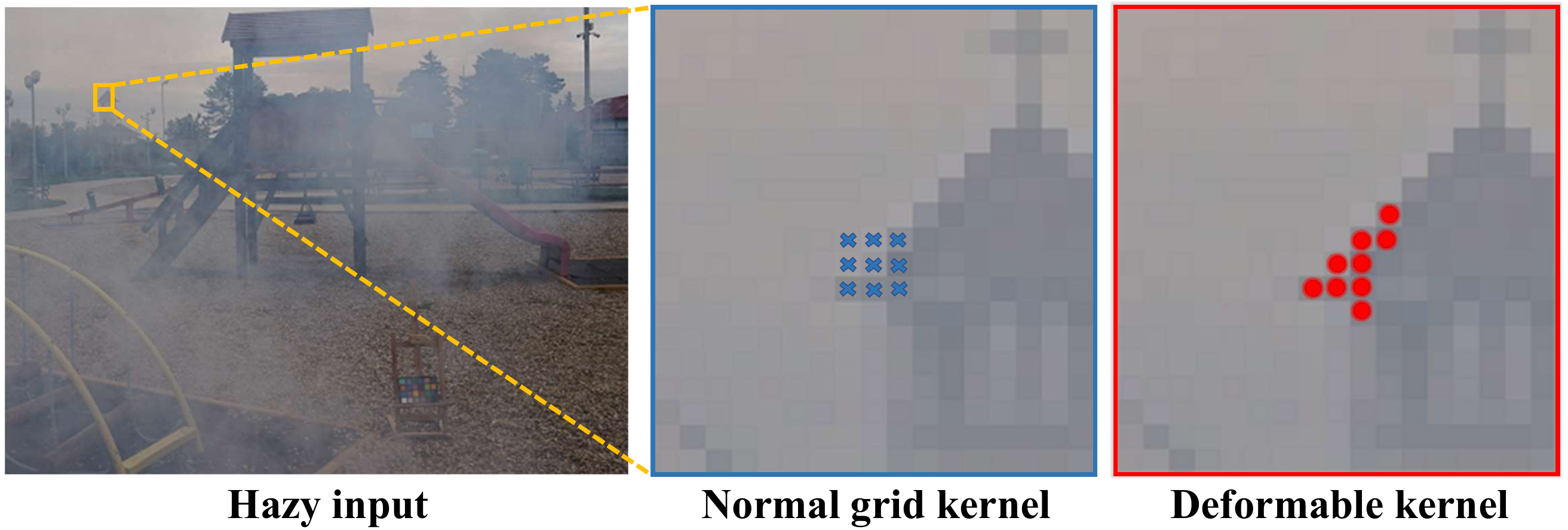}
	\caption{Dynamic feature enhancement module.}
	\label{Fig:DFE}
	\vspace{-0.2cm}
\end{figure}

\subsubsection{Dynamic Feature Enhancement}
Previous works \cite{Qu_2019_CVPR, liuICCV2019GridDehazeNet, qin2020ffa, hong2020distilling, Dong_2020_CVPR, Shao_2020_CVPR} usually employ the fixed grid kernel (\eg 3x3) as shown in Fig. \ref{Fig:DFE} middle, which limits the receptive field and cannot exploit the structured information in the feature space \cite{xu2019learning}. Alternatively, the dilated convolutional layer \cite{yu2015multi} is introduced to expanse the receptive field. However, it will potentially cause the gridding artifacts. 
On the other hand, the shape of receptive field is also important to enlarge the receptive field. As shown in Fig. \ref{Fig:DFE} right, the deformable convolution can capture more important information since the kernel is dynamic and flexible. In fact, the work \cite{xu2019learning} has demonstrated that spatially-invariant convolution kernels could result in corrupted image textures and over-smoothing artifacts, such that the deformable 2D kernels was proposed to enhance the feature for image denoising.
Therefore, we introduce dynamic feature enhancement module (DFE) via deformable convolution \cite{dai2017deformable} to expand receptive field with adaptive shape and improve the model's transformation capability for better image dehazing. 
In particular, we employ two deformable convolutional layers to enable more free-form deformation of the sampling grid, as shown in Fig. \ref{Fig:CDNet}. As such, the network can dynamically pay more attention to the computation of the interest region to fuse more spatially structured information. We also find that DFE deployed after the deep layer achieves better performance than the shallow layers. 

\subsection{Contrastive Regularization}
\label{CR}

Inspired by contrastive learning \cite{hadsell2006dimensionality,oord2018representation,henaff2020data,he2020momentum,chen2020simple}, 
it aims to learn a representation to pull ``positive '' pairs in some metric space and push apart the representation between ``negative'' pairs. 
We propose a new contrastive regularization (CR) to generate better restored images. 
Therefore, we need to consider two aspects in CR: one is to construct the ``positive'' pairs and ``negative'' pairs, the other one is to find the latent feature space of these pairs for contrast. In our CR, the positive pair and negative pair are generated by the group of a clear image $J$ and its restored image $\hat{J}$ by the AE-like dehazing network $\phi$, and the group of $\hat{J}$ and a hazy image $I$, respectively. For simplicity, we call the restored image, the clear image and the hazy image as anchor, positive and negative, respectively. For the latent feature space, we select the common intermediate feature from the same fixed pre-trained model $G$, \eg VGG-19 \cite{simonyan2014very}. 
Thus, the objective function in Eq. (\ref{loss_IR}) can be reformulated as:
\begin{equation}
min\|J-\phi(I,w)\|+\beta\cdot\rho\big(G(I),G(J),G(\phi(I,w))\big),
\label{loss_CR}
\end{equation}
where the first term is the reconstruction loss to align between the restored image and its ground-truth in the data field. We employ L1 loss, as it achieves the better performance compared to L2 loss \cite{zhao2016loss}. The second term $\rho\big(G(I),G(J),G(\phi(I,w))\big)$ is the contrastive regularization among $I,J$ and $\phi(I,w)$ under the same latent feature space, which plays a role of \emph{opposing forces} pulling the restored image $\phi(I,w)$ to its clear image $J$ and pushing $\phi(I,w)$ to its hazy image $I$. $\beta$ is a hyperparameter for balancing the reconstruction loss and CR. To enhance the contrastive ability, we extract the hidden features from different layers of the fixed pre-trained model.

Therefore, the overall dehazing loss function Eq. (\ref{loss_CR}) can be further formulated as:
\begin{equation}
\small
min\|J-\phi(I,w)\|_1+\beta\sum_{i=1}^n\omega_{i}\cdot\frac{D\big(G_i(J), G_i(\phi(I,w))\big)}{D\big(G_i(I), G_i(\phi(I,w))\big)},
\label{loss_overall}
\end{equation}
where $G_i, i=1,2,\cdots n$ extracts the $i$-th hidden features from the fixed pre-trained model. $D(x,y)$ is the L1 distance between $x$ and $y$. $\omega_{i}$ is a weight coefficient. Eq. (\ref{loss_overall}) can be trained via an optimizer (\eg Adam) in an end-to-end manner. Related to our CR, perceptual loss \cite{johnson2016perceptual} measures the visual difference between the prediction and the ground truth by leveraging multi-layer features extracted from a pre-trained deep neural network. Different from the perceptual loss with positive-oriented regularization, we also adopt hazy image (input of dehazing network) as negatives to constrain the solution space, and experiments demonstrate our CR outperforms it for image dehazing (see Section \ref{Effectiveness-CR}). 


\begin{table*}[t]
	\caption{Quantitative comparisons with SOTA methods on the synthetic and real-world dehazing datasets.} 
	\label{performance_SOTS}
	\centering
	\begin{tabular}{c|cc|cc|cc|c}
		\toprule[1pt]
		\multirow{2}{*}{Method} & \multicolumn{2}{c|}{SOTS \cite{li2019benchmarking}}             & \multicolumn{2}{c|}{Dense-Haze \cite{Dense-Haze_2019}}             & \multicolumn{2}{c|}{NH-HAZE \cite{Ancuti_NH-HAZE_2020}}                & \multirow{2}{*}{\# Param} \\ \cline{2-7}
		& \multicolumn{1}{c}{PSNR} & SSIM      & \multicolumn{1}{c}{PSNR} & SSIM            & \multicolumn{1}{c}{PSNR} & SSIM            &                        \\ \hline
		(TPAMI'10) DCP \cite{he2010single}                     & 15.09                     & 0.7649    & 10.06                     & 0.3856          & 10.57                     & 0.5196          & -                      \\
		(TIP'16) DehazeNet \cite{cai2016dehazenet}               & 20.64                     & 0.7995    & 13.84                     & 0.4252          & 16.62                     & 0.5238          & 0.01M                  \\
		(ICCV'17) AOD-Net \cite{li2017aod}                & 19.82                     & 0.8178    & 13.14                     & 0.4144          & 15.40                     & 0.5693          & 0.002M                 \\
		(ICCV'19) GridDehazeNet \cite{liuICCV2019GridDehazeNet}          & 32.16                     & 0.9836    & 13.31                     & 0.3681          & 13.80                     & 0.5370          & 0.96M                  \\
		(AAAI'20) FFA-Net \cite{qin2020ffa}                & 36.39                    & 0.9886   & 14.39                     & 0.4524          & 19.87                     & 0.6915          & 4.68M                  \\
		(CVPR'20) MSBDN \cite{Dong_2020_CVPR}                  & 33.79                     & 0.9840    & 15.37                  & \textbf{0.4858}          & 19.23                     & 0.7056        & 31.35M                 \\
		(CVPR'20) KDDN \cite{hong2020distilling}                   & 34.72                     & 0.9845    & 14.28                     & 0.4074          & 17.39                     & 0.5897          & 5.99M                  \\ 
		(ECCV'20) FDU \cite{dong2020physics} & 32.68& 0.9760& - & -& -& -&- \\ \hline
		Ours         & \textbf{37.17}                 & \textbf{0.9901}  & \textbf{15.80}            & 0.4660 & \textbf{19.88}            & \textbf{0.7173} & 2.61M                  \\ \bottomrule[1pt]
	\end{tabular}
	\vspace{-0.3cm}
\end{table*}

\begin{figure*}[t]
\centering
 \subfigure[Hazy input]{
		\includegraphics[width = 0.17\linewidth]{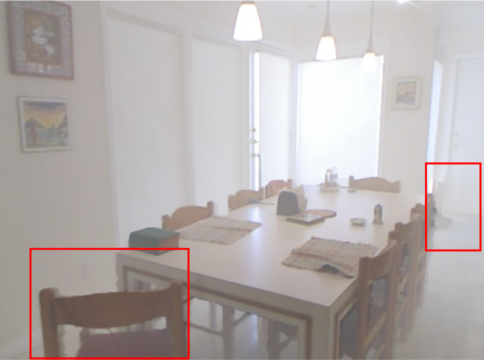}
		\label{Fig:SOTS_a}
	}
	\subfigure[DCP \cite{he2010single}]{
		\includegraphics[width = 0.17\linewidth]{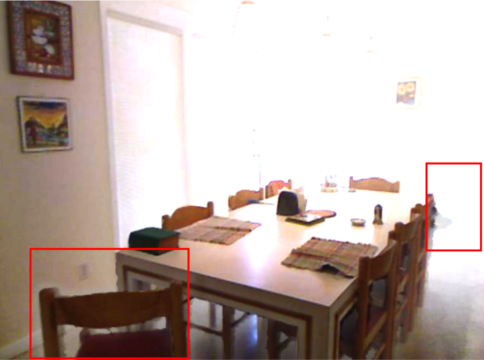}
		\label{Fig:SOTS_b}
	}
	\subfigure[DehazeNet \cite{cai2016dehazenet}]{
		\includegraphics[width = 0.17\linewidth]{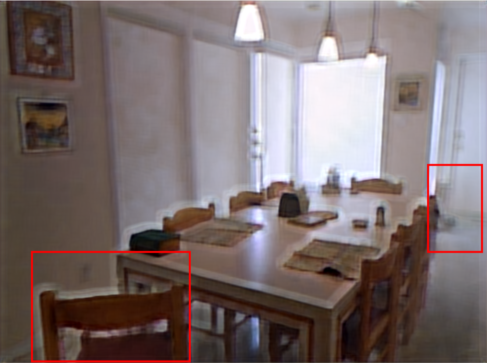}
		\label{Fig:SOTS_c}
	}
	\subfigure[AOD-Net \cite{li2017aod}]{
		\includegraphics[width = 0.17\linewidth]{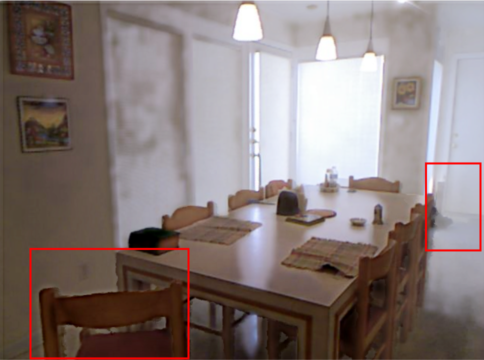}
		\label{Fig:SOTS_d}
	}
	\vspace{-0.8em}
	\subfigure[GridDehazeNet \cite{liuICCV2019GridDehazeNet}]{
		\includegraphics[width = 0.17\linewidth]{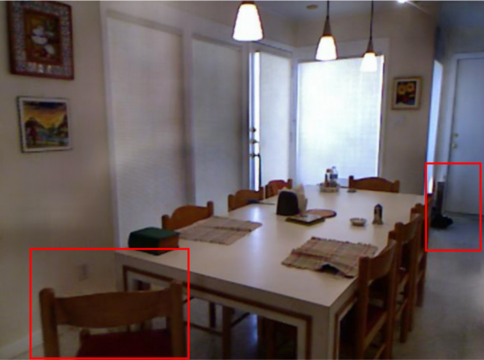}
		\label{Fig:SOTS_e}
	}
	\vspace{-0.2cm}
	\subfigure[FFA-Net \cite{qin2020ffa}]{
		\includegraphics[width = 0.17\linewidth]{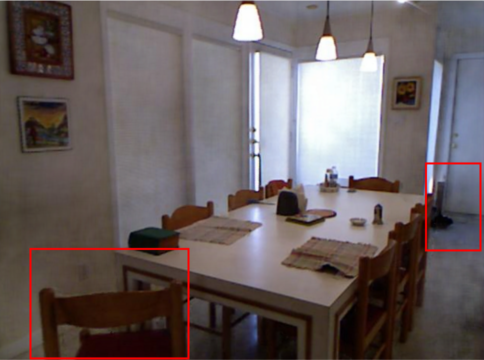}
		\label{Fig:SOTS_f}
	}
	\subfigure[MSBDN \cite{Dong_2020_CVPR}]{
		\includegraphics[width = 0.17\linewidth]{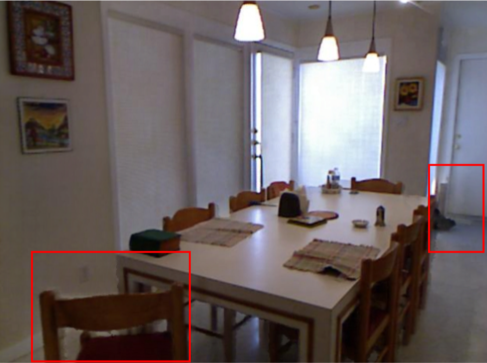}
		\label{Fig:SOTS_g}
	}
	\subfigure[KDDN \cite{hong2020distilling}]{
		\includegraphics[width = 0.17\linewidth]{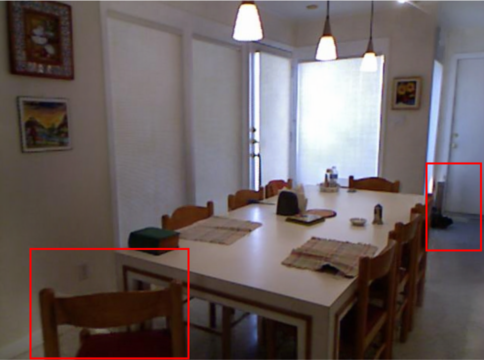}
		\label{Fig:SOTS_h}
	}
	\subfigure[Ours]{
		\includegraphics[width = 0.17\linewidth]{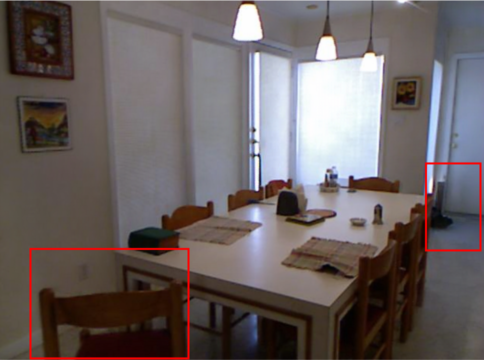}
		\label{Fig:SOTS_i}
	}
	\subfigure[Ground-truth]{
		\includegraphics[width = 0.17\linewidth]{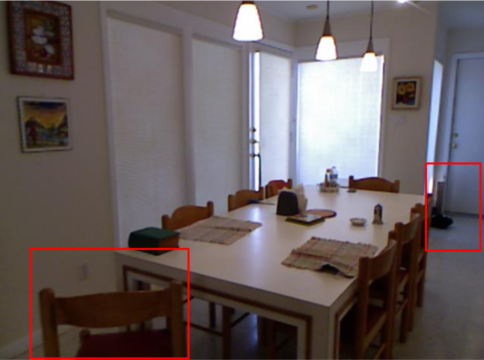}
		\label{Fig:SOTS_j}
	}
	\vspace{0.1cm}
 \caption{Visual results comparison on SOTS \cite{li2019benchmarking} dataset.  Zoom in for best view.}
 \label{Fig:SOTS}
\vspace{-1em}
\end{figure*}

\section{Experiments}
\label{Experiments}

\subsection{Experiment Setup}

\textbf{Implementation Details.}
Our AECR-Net is implemented by PyTorch 1.2.0 and MindSpore with one NVIDIA TITAN RTX GPU. The models are trained using Adam optimizer with exponential decay rates $\beta_{1}$ and $\beta_{2}$ equal to 0.9 and 0.999, respectively. The initial learning rate and batch-size are set to 0.0002 and 16, respectively. We use cosine annealing strategy \cite{he2019bag} to adjust the learning rate. We empirically set the penalty parameter $\beta$ to 0.1 and the total number of epoch to 100. We set the L1 distance loss in Eq. (\ref{loss_overall}) after the latent features of the $1$st, $3$rd, $5$th, $9$th and $13$th layers from the fixed pre-trained VGG-19, and their corresponding coefficients $\omega_{i}, i=1,\cdots,5$ to $\frac{1}{32}$, $\frac{1}{16}$, $\frac{1}{8}$, $\frac{1}{4}$ and $1$, respectively.

\textbf{Datasets.}
We evaluate the proposed method on synthetic dataset and real-world datasets. RESIDE \cite{li2019benchmarking} is a widely used synthetic dataset, which consists of five subsets: Indoor Training Set (ITS), Outdoor Training Set (OTS), Synthetic Objective Testing Set (SOTS), Real World task-driven Testing Set (RTTS), and Hybrid Subjective Testing Set (HSTS). ITS, OTS and SOTS are synthetic datasets, RTTS is the real-world dataset, HSTS consists of synthetic and real-word hazy images. Following the works \cite{liuICCV2019GridDehazeNet,qin2020ffa,hong2020distilling,Dong_2020_CVPR}, we select ITS and SOTS indoor as our training and testing datasets. In order to further evaluate the robustness of our method in the real-world scene, we also adopt two real-world datasets: Dense-Haze \cite{Dense-Haze_2019} and NH-HAZE \cite{Ancuti_NH-HAZE_2020}. More details are provided in the supplementary.

\textbf{Evaluation Metric and Compatitors.}
To evaluate the performance of our method, we adopt the Peak Signal to Noise Ratio (PSNR) and the Structural Similarity index (SSIM) as the evaluation metrics, which are usually used as criteria to evaluate image quality in the image dehazing task. We compare with the prior-based method (\eg DCP \cite{he2010single}), physical model based methods (\eg DehazeNet \cite{cai2016dehazenet} and AOD-Net \cite{li2017aod}), and hazy-to-clear image translation based methods (\eg GridDehazeNet \cite{liuICCV2019GridDehazeNet}, FFA-Net \cite{qin2020ffa}, MSBDN \cite{Dong_2020_CVPR} and KDDN \cite{hong2020distilling}).

\subsection{Comparison with State-of-the-art Methods}
\textbf{Results on Synthetic Dataset.}
In Table \ref{performance_SOTS}, we summarize the performance of our AECR-Net and SOTA methods on RESIDE dataset \cite{li2019benchmarking} (\emph{a.k.a}, SOTS). Our AECR-Net achieves the best performance with 37.17dB PSNR and 0.9901 SSIM, compared to SOTA methods. In particular, compared to FFA-Net \cite{qin2020ffa} with the second top performance, our AECR-Net achieves 0.78dB PSNR and 0.0015 SSIM performance gains with the significant reduction of 2M parameters. We also compare our AECR-Net with SOTA methods on the quality of the restored images, which is shown in Fig. \ref{Fig:SOTS}. We can observe that DCP \cite{he2010single} and  DehazeNet \cite{cai2016dehazenet} and AOD-Net \cite{li2017aod} cannot successfully remove dense haze, and suffer from the color distortion (see Fig. \ref{Fig:SOTS_b}-\ref{Fig:SOTS_d}). Compared to DCP, DehazeNet and AOD-Net, the hazy-to-clear image translation based methods in an end-to-end manner (\eg GridDehazeNet \cite{liuICCV2019GridDehazeNet}, FFA-Net \cite{qin2020ffa}, MSBDN \cite{Dong_2020_CVPR} and KDDN \cite{hong2020distilling}) achieve the restored images with higher quality. However, they still generate some gray mottled artifacts as shown in Fig. \ref{Fig:SOTS_e}-\ref{Fig:SOTS_f} and cannot completely remove the haze in some regions (see the red rectangles of Fig. \ref{Fig:SOTS_g}-\ref{Fig:SOTS_h}). Our method generates the most natural images and achieves the similar patterns to the ground-truth both in low and high frequency regions. More examples can be found in the supplementary.


\begin{figure*}[t]
\centering
  \subfigure[Hazy input]{
    \includegraphics[width = 0.17\linewidth]{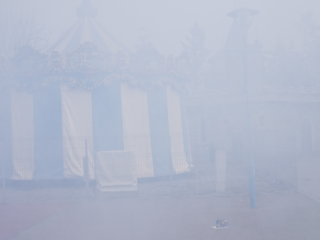}
    \label{Fig:DH_a}
    }
  \subfigure[DCP \cite{he2010single}]{
    \includegraphics[width = 0.17\linewidth]{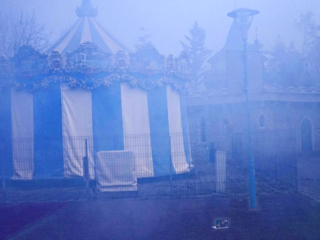}
    \label{Fig:DH_b}
  }
  \subfigure[DehazeNet \cite{cai2016dehazenet}]{
    \includegraphics[width = 0.17\linewidth]{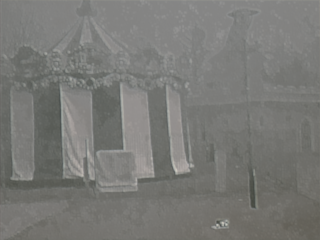}
    \label{Fig:DH_c}
  }
  \subfigure[AOD-Net \cite{li2017aod}]{
    \includegraphics[width = 0.17\linewidth]{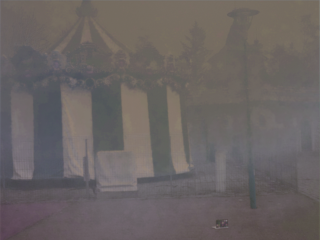}
    \label{Fig:DH_d}
  }
  \vspace{-0.8em}
  \subfigure[GridDehazeNet \cite{liuICCV2019GridDehazeNet}]{
    \includegraphics[width = 0.17\linewidth]{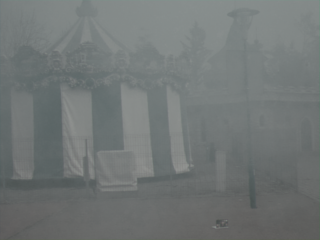}
    \label{Fig:DH_e}
  }
  
  \subfigure[FFA-Net \cite{qin2020ffa}]{
    \includegraphics[width = 0.17\linewidth]{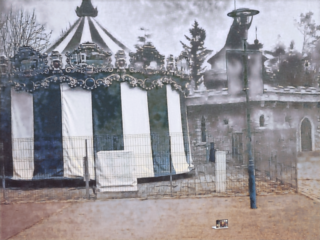}
    \label{Fig:DH_f}
  }
  \subfigure[MSBDN \cite{Dong_2020_CVPR}]{
    \includegraphics[width = 0.17\linewidth]{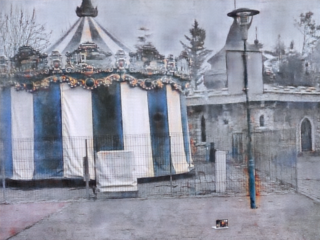}
    \label{Fig:DH_g}
  }
  \subfigure[KDDN \cite{hong2020distilling}]{
    \includegraphics[width = 0.17\linewidth]{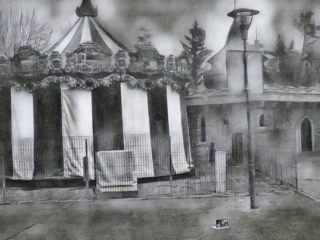}
    \label{Fig:DH_h}
  }
  \subfigure[Ours]{
    \includegraphics[width = 0.17\linewidth]{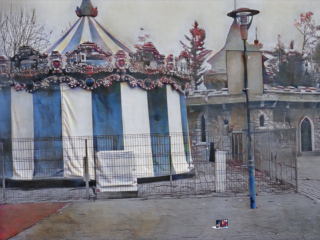}
    \label{Fig:DH_i}
  }
  \subfigure[Ground-truth]{
    \includegraphics[width = 0.17\linewidth]{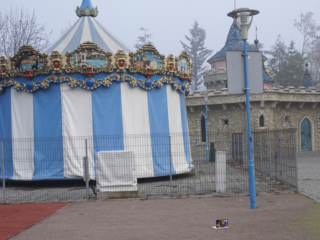}
    \label{Fig:DH_j}
  }
 \caption{Visual comparison on the Dense-Haze dataset.}
 \label{Fig:DH}
\vspace{-1em}
\end{figure*}


\begin{figure*}[t]
\centering
  \subfigure[Hazy input]{
    \includegraphics[width = 0.17\linewidth]{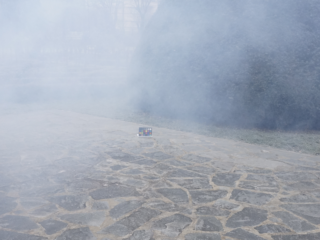}
    \label{Fig:NH_a}
    }
  \subfigure[DCP \cite{he2010single}]{
    \includegraphics[width = 0.17\linewidth]{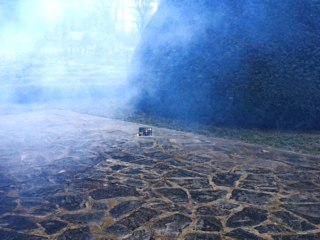}
    \label{Fig:NH_b}
  }
  \subfigure[DehazeNet \cite{cai2016dehazenet}]{
    \includegraphics[width = 0.17\linewidth]{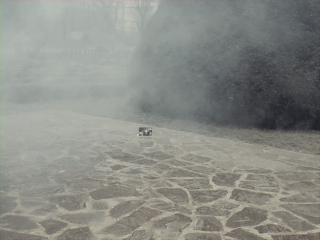}
    \label{Fig:NH_c}
  }
  \subfigure[AOD-Net \cite{li2017aod}]{
    \includegraphics[width = 0.17\linewidth]{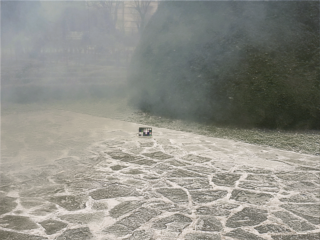}
    \label{Fig:NH_d}
  }
  \vspace{-0.8em}
  \subfigure[GridDehazeNet \cite{liuICCV2019GridDehazeNet}]{
    \includegraphics[width = 0.17\linewidth]{44_GridDehazeNet.png}
    \label{Fig:NH_e}
  }
  \subfigure[FFA-Net \cite{qin2020ffa}]{
    \includegraphics[width = 0.17\linewidth]{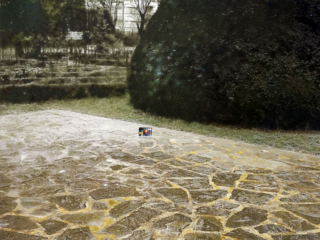}
    \label{Fig:NH_f}
  }
  \subfigure[MSBDN \cite{Dong_2020_CVPR}]{
    \includegraphics[width = 0.17\linewidth]{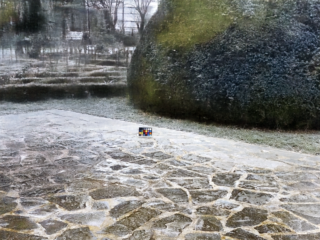}
    \label{Fig:NH_g}
  }
  \subfigure[KDDN \cite{hong2020distilling}]{
    \includegraphics[width = 0.17\linewidth]{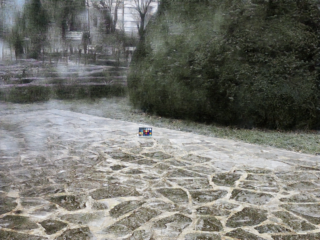}
    \label{Fig:NH_h}
  }
  \subfigure[Ours]{
    \includegraphics[width = 0.17\linewidth]{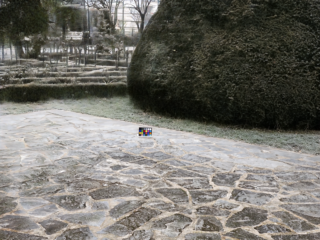}
    \label{Fig:NH_i}
  }
  \subfigure[Ground-truth]{
    \includegraphics[width = 0.17\linewidth]{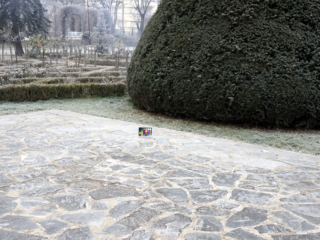}
    \label{Fig:NH_j}
  }
 \caption{Visual comparison on NH-HAZE datasets.} 
\label{Fig:NH}
\vspace{-1em}
\end{figure*}

\textbf{Results on Real-world Datasets.}
We also compare our AECR-Net with SOTA methods on Dense-Haze \cite{Dense-Haze_2019} and NH-HAZE \cite{Ancuti_NH-HAZE_2020} datasets.
As shown in Table \ref{performance_SOTS}, we can observe: (1) Our AECR-Net outperforms all SOTA methods with 19.88dB PSNR and 0.7173 SSIM on NH-HAZE dataset. 
(2) Our AECR-Net also achieves the highest PSNR of 15.80dB, compared to SOTA methods. Note that MSBDN achieves only about 0.02 higher SSIM, but with 12$\times$ parameters, compared to our AECR-Net. %
(3) Compared to RESIDE dataset, Dense-Haze and NH-HAZE dataset are more difficult to remove the haze, especially on Dense-Haze dataset. This is due to the real dense haze which leads to the severe degradation of information. 
We also compare our AECR-Net with SOTA methods on the quality of restored images,  which are presented in Fig. \ref{Fig:DH} and Fig. \ref{Fig:NH}. 
Obviously, our AECR-Net generates the most natural images, compared to other methods. The restored images by DCP \cite{he2010single}, DehazeNet \cite{cai2016dehazenet}, AOD-Net \cite{li2017aod}, GridDehazeNet \cite{liuICCV2019GridDehazeNet}, FFA-Net \cite{qin2020ffa} and KDDN \cite{hong2020distilling} suffer from the serious color distortion and texture loss. Besides, there are still some thick haze existed in the restored images by MSBDN \cite{Dong_2020_CVPR} and KDDN \cite{hong2020distilling}. More examples can be found in the supplementary.

\begin{table}[t]
	\caption{Ablation study on AECR-Net. * denotes only positive samples are used for training. SC means skip connection.}
	\label{ablation_netwrok}
	\footnotesize
	\centering
	\begin{tabular}{cc|cc}
		\toprule[1pt]
		Model & CR& PSNR& SSIM \\
		\hline 
		base& -& 33.85& 0.9820\\
		base+mixup & -& 34.04& 0.9838\\
		base+DFE & -& 35.50& 0.9853\\
		base+DFE+SC & - & 35.59& 0.9858\\
		base+DFE+mixup & -& 36.20& 0.9869\\ \hline
		base+DFE+mixup+CR* & $\surd$(w/o negative)& 36.46& 0.9889\\
		Ours& $\surd$& \textbf{37.17}& \textbf{0.9901}\\
		\bottomrule[1pt]
	\end{tabular}
\vspace{-0.3cm}
\end{table}

\subsection{Ablation Study}
\label{ablation_study}
To demonstrate the effectiveness of the proposed AECR-Net, we conduct ablation study to analyze different elements, including mixup, DFE and CR. 

We first construct our \emph{base} network as the baseline of dehazing network, which mainly consists of two downsampling layers, six FA blocks and two upsampling layers.
Subsequently, we add the different modules into base network as: (1) \textbf{base+mixup}: Add the mixup operation into baseline. (2) \textbf{base+DFE}: Add the DFE module into baseline. (3) \textbf{base+DFE+mixup}: Add both DFE module and mixup operation into baseline, \emph{a.k.a.} our AE-like dehazing network.
(4) \textbf{base+DFE+mixup+CR*}: Add CR without using negative samples into our AE-like dehazing network. It means that only positive samples are utilized to train the dehazing network.
(5) \textbf{Ours}: The combination of our AE-like dehazing network and the proposed CR, which allows both negative and positive samples for training.

We employ L1 loss as image reconstruction loss (\ie the first term in Eq. (\ref{loss_overall})), and use RESIDE \cite{li2019benchmarking} dataset for both training and testing. The performance of these models are summarized in Table \ref{ablation_netwrok}.

\textbf{Effect of Adaptive Mixup Operation.}
Adaptive mixup operation can improve the dehazing network with additional negligible parameter, which provides additional flexibility to fuse the different features. In Table \ref{ablation_netwrok}, it can improve the performance of our base network, \eg the increases of 0.19dB and 0.7dB in PSNR from base to base+mixup and from base+DFE to base+DFE+mixup, respectively.
Furthermore, we compare our adaptive mixup operation with skip connection (SC) operation. The factors (\ie $\sigma(\theta_1)$ and $\sigma(\theta_2)$ in Eq. (\ref{skip_connection})) in our adaptive mixup operation are learnable, while SC has the identical information fusion. Adaptive mixup operation achieves 0.61dB PSNR gains over SC.

\textbf{Effect of DFE Module.}
\label{exp:DFE}
DFE module significantly improves the performance from base to base+DFE with an increase of 1.65dB PSNR and from base+mixup to base+DFE+mixup with an increase of 2.16dB PSNR. Therefore, DFE is an more important factor than adaptive mixup, due to the higher performance gains. 
We also evaluate the effect of DFE positions before and after 6 FA blocks. The results demonstrate that  
DFE deployed after the deeper layers achieves better performance than the shallow layers. The detailed performance are shown in the supplementary.

\textbf{Effect of Contrastive Regularization.}
\label{Effectiveness-CR}

We consider the effect of CR whether uses negative samples. CR* represents only positive samples are used for training, which is similar to perceptual loss \cite{johnson2016perceptual}. Compared to base+DFE+mixup, adding CR* on that (\ie base+DFE+mixup+CR*) only achieves slightly higher PSNR and SSIM with the gains of 0.26dB and 0.002, respectively. Our AECR-Net employs the proposed CR adding both negative and positive samples for training, which significantly achieves performance gains over base+DFE+mixup+CR*. For example, our AECR-Net achieves a higher PSNR of 37.17dB, compared to base+DFE+mixup+CR* with 36.46dB PSNR.

\subsection{Universal Contrastive Regularization}
To evaluate the universality of the proposed CR, we add our CR into various SOTA methods \cite{liuICCV2019GridDehazeNet, qin2020ffa, Dong_2020_CVPR,hong2020distilling}. As presented in Table \ref{plug-and-play_ablation}, CR can further improve the performance of SOTA methods. In other words, our CR is model-agnostic to train the dehazing networks effectively. Furthermore, our CR cannot increase the additional parameters for inference, since it can be directly removed for testing.

CR can also enhance the visual quality of SOTA methods. For example, adding our CR into SOTA methods can reduce the effect of black spots and color distortion (see supplementary on these examples). 

\begin{table}[th]
	\caption{Results of applying CR into SOTA methods.}
	\label{plug-and-play_ablation}
	\small
	\centering
	\begin{tabular}{c|cc}
		\toprule[1pt]
		Method & PSNR& SSIM \\
		\hline 
		GridDehazeNet \cite{liuICCV2019GridDehazeNet} & 32.99 (\textcolor{red}{$\uparrow$} 0.83)
		& 0.9863 (\textcolor{red}{$\uparrow$} 0.0027)\\
		FFA-Net \cite{qin2020ffa}& 36.74 (\textcolor{red}{$\uparrow$} 0.35)& 0.9906 (\textcolor{red}{$\uparrow$} 0.0020)\\
		KDDN \cite{hong2020distilling}& 35.18 (\textcolor{red}{$\uparrow$} 0.46)& 0.9854 (\textcolor{red}{$\uparrow$} 0.0009) \\
		MSBDN \cite{Dong_2020_CVPR}& 34.45 (\textcolor{red}{$\uparrow$} 0.66)& 0.9861 (\textcolor{red}{$\uparrow$} 0.0021) \\
		\bottomrule[1pt]
		
	\end{tabular}
	\vspace{-0.2cm}
\end{table}
 
\subsection{Discussion}
We further explore the effect of different rates (\ie $r$) between positive and negative samples on CR. 
If the number of negative samples is $r$, we will take the current hazy input as one sample and randomly select the other $r-1$ negative samples from the the same batch to the input haze image.  
For positive samples, we select the corresponding clear images to the selected negative samples as positive ones. We select our AECR-Net with the rate of 1:1 as baseline, and conduct all experiments on RESIDE dataset. Additionally, we consider at most 10 positive or negative samples, because of the limited GPU memory size.

As shown in Table \ref{multiple_contrastive},
adding more negative samples into CR achieves the better performance, while adding more positive samples achieves the opposite results. We conjecture this is due to the different positive pattern that confuses the anchor to learn good pattern. For negative samples, the more negative samples, the farther away from the worse pattern in the hazy images. Therefore, our AECR-Net with the rate of 1:10 achieves the best performance. However, it takes longer training time when increasing the number of negative samples. For example, Our AECR-Net with the rate of 1:10 takes about 200 hours in total (\ie 2$\times$) for training, compare to  total 100 hours at the rate of 1:1\footnote{All tables and figures report the results
of the rate 1:1, except Table \ref{multiple_contrastive}.}.    

\begin{table}[t]
	\caption{Comparisons of different positive and negative sample rates on CR. The baseline is AECR-Net with the rate of 1:1.}
	\label{multiple_contrastive}
	\centering
	\begin{tabular}{ccc|cc} 
		\toprule[1pt]
		Rate & \# Positive & \# Negative & PSNR & SSIM\\
		\hline
		1:1 & 1 & 1 & 37.17& 0.9901\\
		1:$r$ & 1 & 10 &   \textbf{37.41}&  \textbf{0.9906}\\
		$r$:1 & 10 & 1 &  35.61& 0.9862 \\
		$r$:$r$ & 10 & 10 &  35.65& 0.9861 \\
		\bottomrule[1pt]
	\end{tabular}
	\vspace{-0.3cm}
\end{table}

\section{Conclusion}
In this paper, we propose a novel AECR-Net for single image dehazing, which consists of contrastive regularization (CR) and autoencoder-like (AE) network. %
CR is built upon contrastive learning to ensure that the restored image is pulled to closer to the clear image and pushed to far away from the hazy image in representation space. 
AE-like dehazing network based on the adaptive mixup operation and a dynamic feature enhancement module is compact and benefits from preserving information flow adaptively and expanding the receptive field to improve the network's transformation capability. We have comprehensively evaluated the performance of AECR-Net on synthetic and real-world datasets, which demonstrates the superior performance gains over the SOTA methods. 

\section{Acknowledgements}
This work is supported by the National Natural Science Foundation of China 61772524, 61876161, 61972157; the National Key Research and Development Program of China No.2020AAA0108301; Natural Science Foundation of Shanghai (20ZR1417700); CAAI-Huawei MindSpore Open Fund; the Research Program of Zhejiang Lab (No.2019KD0AC02).

{\small
\bibliographystyle{ieee_fullname}
\bibliography{egbib}
}

\end{document}